\documentclass{article}
\usepackage{spconf,amsmath,graphicx}
\usepackage{url}            
\usepackage{booktabs}       
\usepackage{amsfonts}       
\usepackage{nicefrac}       
\usepackage{microtype}      
\usepackage{xcolor}         
\usepackage{graphicx}
\usepackage{multirow}
\usepackage{bbding}
\usepackage{enumitem}
\usepackage[colorlinks,linkcolor=black,anchorcolor=black,citecolor=black]{hyperref}

\let\OLDthebibliography\thebibliography
\renewcommand\thebibliography[1]{
  \OLDthebibliography{#1}
  \setlength{\parskip}{0pt}
  \setlength{\itemsep}{0pt plus 0.3ex}
}

\begin{document}\sloppy

\def\x{{\mathbf x}}
\def\L{{\cal L}}

\title{High-Fidelity Lake Extraction via Two-Stage Prompt Enhancement:\\Establishing a Novel Baseline and Benchmark}
%
\name{Ben Chen$^{1, }$$^*$, Xuechao Zou$^{1, }$$^*$, Kai Li$^{2}$, Yu Zhang$^{1}$, Junliang Xing$^{2}$, Pin Tao$^{1,2, }$$^\dagger$\thanks{$^*$Equal contribution.}\thanks{$^\dagger$Corresponding author.}}
\address{
$^1$Qinghai University, Department of Computer Technology and Applications, Xining, China\\
$^2$Tsinghua University, Department of Computer Science  and Technology, Beijing, China
}

\maketitle

\begin{abstract}
Lake extraction from remote sensing imagery is a complex challenge due to the varied lake shapes and data noise. Current methods rely on multispectral image datasets, making it challenging to learn lake features accurately from pixel arrangements. This, in turn, affects model learning and the creation of accurate segmentation masks. This paper introduces a prompt-based dataset construction approach that provides approximate lake locations using point, box, and mask prompts. We also propose a two-stage prompt enhancement framework, LEPrompter, with prompt-based and prompt-free stages during training. The prompt-based stage employs a prompt encoder to extract prior information, integrating prompt tokens and image embedding through self- and cross-attention in the prompt decoder. Prompts are deactivated to ensure independence during inference, enabling automated lake extraction without introducing additional parameters and GFlops. Extensive experiments showcase performance improvements of our proposed approach compared to the previous state-of-the-art method. The source code is available at \url{https://github.com/BastianChen/LEPrompter}.
\end{abstract}
\section{Introduction}
\label{sec:intro}

Lakes are important environmental and climatic indicators and have received significant scientific attention~\cite{ClimaticIndicator}. Global observation technology and sensing devices have made remote sensing imagery a popular tool for extracting water bodies~\cite{HA-Net}. Automated lake extraction from remote sensing imagery is crucial in monitoring climate change~\cite{MSCANet}.


\begin{figure}[t]
\centering
\includegraphics[width=0.9\linewidth]{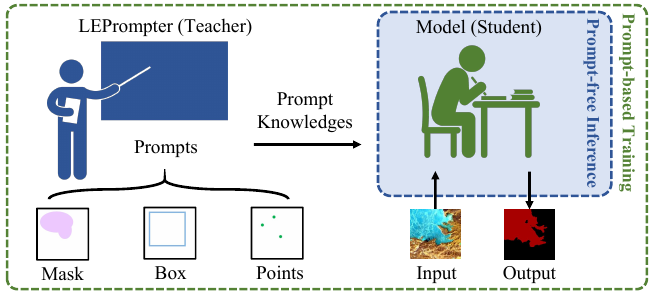}
\caption{Our proposed two-stage prompt enhancement framework for lake extraction. The prompt-based approach simulates a teacher guiding students to solve challenging problems, while the prompt-free approach allows students to tackle problems independently. Conversely, the inference process exclusively utilizes the prompt-free approach.}
\label{TeacherAndStudernt}
\end{figure}

Lake extraction is a semantic segmentation task that researchers have recently tackled using deep learning techniques, such as UDGN~\cite{UDGN}, to enhance the spatial resolution of lake regions. MSLWENet~\cite{QTPL-Dataset} proposes an end-to-end multi-scale plateau lake extraction network model based on ResNet-101~\cite{ResNet} and depth-wise separable convolution~\cite{Depth-wiseConvolutionLayer}. LEFormer~\cite{leformer} utilizes a hybrid CNN-Transformer architecture, HA-Net~\cite{HA-Net} employs a mixed-scale attention model, and MSNANet~\cite{MSNANet} adopts an encoder-decoder framework to improve feature representation and extraction results. However, existing lake datasets~\cite{QTPL-Dataset} consist of multi-channel input images with single-channel ground truth, posing challenges for model learning due to the complex pixel information and potential noise. Models need to capture intricate lake details effectively and handle noise to improve extraction accuracy. 

Recently, the prompt-based interactive semantic segmentation method SAM~\cite{SAM} has shown promising performance by refining predicted masks through prompt prior information. However, this method requires a pre-prompt (point, box, or mask) with the input image, making it interactive. Additionally, SAM's segmentation results lack semantic information, which restricts their suitability for fully automated interpretation of remote sensing images. Furthermore, SAM utilizes extensive backbones, such as ViT-H~\cite{ViT}, leading to numerous model parameters and increased computational complexity.

To address the limitations of existing datasets and methods, we employ morphological operations to create a prompt dataset based on the ground truth, which consists of point, box, and mask prompts. We propose LEPrompter, a two-stage prompt enhancement framework for lake extraction during training, to leverage the prompt dataset effectively. The framework includes prompt-based and prompt-free stages. Training progresses in the prompt-based stage until reaching a specific step threshold. After that, we transition to the prompt-free stage, training the lake extraction model independently of prompts. In the prompt-based stage, the prompt dataset is processed by a lightweight prompt encoder and decoder, increasing only 1.23M parameters and 0.95G Flops. This processing is fused with the image embedding obtained from the lake extraction model, which collectively generates output tokens for mask prediction. In inference, we employ the prompt-free approach. This approach requires neither additional model parameters nor GFlops. The workflow is illustrated in Fig.~\ref{TeacherAndStudernt}. Experimental results demonstrate that our proposed approach consistently improves the performance of the previous SOTA methods on the Surface Water dataset (SW dataset) and Qinghai-Tibet Plateau Lake dataset (QTPL dataset)~\cite{QTPL-Dataset}, achieving mIoU of 91.53\% and 97.44\%, respectively. The main contributions of our approach are summarized as follows:

\begin{itemize}
    \item We develop a unified morphological method for generating various prompts and establish three types of prompt datasets (point, box, and mask), serving as a benchmark for lake extraction from remote sensing imagery.
    \item We propose a two-stage prompt enhancement framework for automated lake extraction. This framework employs prompts to guide model training while allowing independence from prompts during inference, serving as a baseline for lake extraction with prompt datasets.
    \item We evaluate the influence of prompt types for lake extraction and observe that a slight prompt positively guides model learning. Conversely, excessive prompts restrict performance improvement, aligning with the scenario of teachers guiding students in the real world.
\end{itemize}


\begin{figure}[t]
\centering
\includegraphics[width=0.4\textwidth]{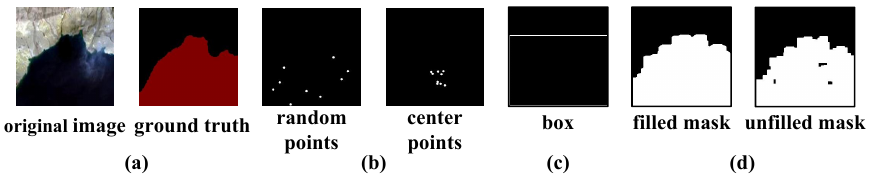} 
\caption{Visualization images of our proposed benchmark.}
\label{Prompt_Datasets}
\end{figure}

\begin{figure*}[t]
\centering
\includegraphics[width=0.73\textwidth]{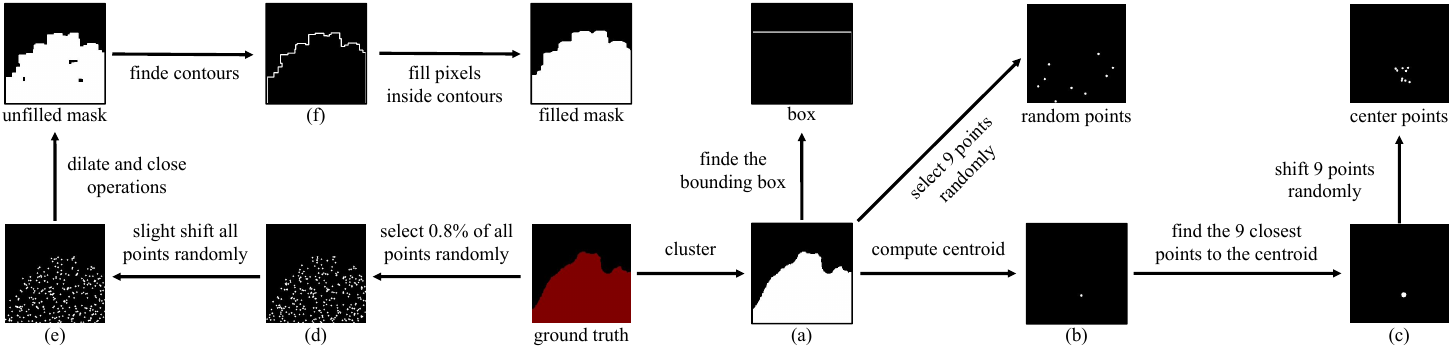} 
\caption{The workflow diagram of creating our benchmark for lake extraction.}
\label{DatasetWorkflowDiagram}
\end{figure*}

\section{Related Work}

\subsection{Lake Extraction} 

Lake extraction aims to automatically locate the boundaries of a lake's location from remote sensing imagery, and it belongs to the semantic segmentation task. Deep learning-based approaches for lake extraction have recently garnered significant attention from researchers. These approaches, such as UDGN~\cite{UDGN}, aim to enhance the spatial resolution of lake regions, but struggle with diverse lake types and spatial-spectral characteristics, leading to lost spatial information. To address these challenges, researchers explore integration strategies to optimize network structures and leverage multi-scale features. For instance, MSLWENet~\cite{QTPL-Dataset} proposes an end-to-end multi-scale plateau lake extraction network model based on ResNet-101~\cite{ResNet} and depth-separable convolution~\cite{Depth-wiseConvolutionLayer}. However, this model is susceptible to noise, particularly for lakes with intricate surface textures. Various approaches have been developed to overcome these challenges in the field of lake extraction. LEFormer~\cite{leformer} utilizes a hybrid CNN-Transformer architecture, leveraging CNN to extract local features and the Transformer to capture global features. HA-Net~\cite{HA-Net} employs a mixed-scale attention model, and MSNANet~\cite{MSNANet} adopts an encoder-decoder framework to enhance feature representation and achieve improved extraction results for lakes. However, the extraction of lakes remains challenging due to high interclass heterogeneity and complex background information.

\subsection{Prompt Learning} 

Prompt learning aims to reduce semantic differences, bridge the pre-training and fine-tuning gap, and prevent overfitting. Since the introduction of GPT-3~\cite{GPT3}, prompt learning has advanced from traditional discrete and continuous prompt construction to large-scale model-oriented in-context learning~\cite{Flamingo}, instruction-tuning~\cite{VisualInstructionTuning}, and chain-of-thought approaches~\cite{Chain-of-thought-prompting}. Current methods for constructing prompts mainly involve manual templates, heuristic-based templates, generation, word embedding fine-tuning, and pseudo tokens. Recently, prompt-based interactive semantic segmentation method SAM~\cite{SAM} has shown remarkable performance. However, its applicability for automatic image segmentation is limited due to the requirement of pre-prompt input and the use of an extensive backbone~\cite{ViT}, resulting in increased model parameters and computational costs. In remote sensing imagery, RSPrompter~\cite{RSPrompter} has successfully introduced prompt mechanisms. However, this method requires additional model structures to extract prompt information during inference, introducing additional parameters and computational costs.

In this work, we generate images containing prompt information based on the ground truth to assist in supervised training with the original dataset. We adopt a prompt-free approach during inference, using only the original lake extraction model without additional parameters or computational costs.

\begin{figure*}[t]
\centering
\includegraphics[width=0.78\linewidth]{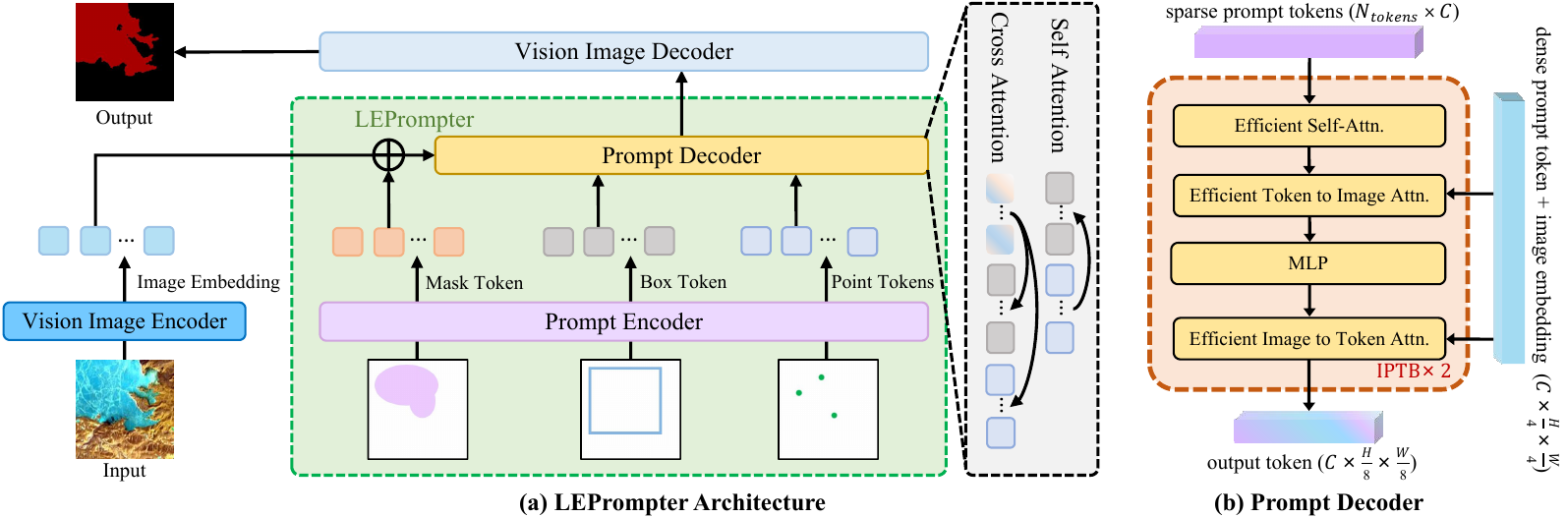} 
\caption{Overview architecture of our proposed enhancement framework LEPrompter and lightweight prompt decoder.}
\label{Overall_architecture}
\end{figure*}


\section{Prompt-based Lake Extraction Dataset}

We evaluate the accuracy of two satellite remote sensing datasets for lake extraction: the SW and QTPL~\cite{QTPL-Dataset} datasets. Both datasets include annotated lake water bodies and have a bit depth of 24 bits per pixel, including R, G, and B spectral bands. The SW dataset contains 17,596 images of size 256 × 256, divided randomly into training and testing sets with a 4:1 ratio. In comparison, the QTPL dataset includes 6,773 images of size 256 × 256, with a spatial resolution of 17 meters, divided randomly into training and testing sets with a 9:1 ratio. However, these datasets suffer from the rich spatial-spectral characteristics of lakes, leading to the loss of crucial spatial information and noise factors that impact the model's learning process. Moreover, these datasets present challenges of high interclass heterogeneity and complex background information, such as snow, glaciers, and mountains, introducing contextual ambiguity and additional extraction challenges.

Previous approaches have focused on improving the model, while our approach directly improves the dataset to reduce the model's learning difficulty. We utilize the density-based clustering algorithm DBSCAN~\cite{DBSCAN} and simple morphological operations, such as erosion and dilation, to refine the ground truth. This process creates our benchmark as a supplementary dataset, called the prompt dataset. In contrast to the interactive semantic segmentation model SAM~\cite{SAM}, our benchmark is generated directly from the ground truth by simulating human prompt habits, including point, box, and mask prompts. Our benchmark comprises five prompt types, matching the number of images in the original dataset's training set. We generate random and center points concentrated in the lake's central area for point prompts, with a maximum of 9 points per type. Masks consist of filled masks with complete interiors and unfilled masks with partially unfilled interiors. The box prompt encompasses the entire lake. An example of our benchmark is shown in Fig. \ref{Prompt_Datasets}, and the workflow diagram for creating our benchmark is depicted in Fig. \ref{DatasetWorkflowDiagram}. The creation process for each prompt is summarized as follows:
\begin{itemize}
    \item By applying DBSCAN clustering, we obtain the pixel map (a) of the lakes and randomly select 9 points as random points prompt. We calculate the centroid of the pixels to generate an image (b) and identify the 9 points closest to the centroid, resulting in the image (c). Finally, a slight shift is applied to the positions of these 9 points to obtain the final center points prompt.
    \item We select the top-left and bottom-right points on the pixel map (a), generated through DBSCAN clustering, to accurately determine the box prompt.
    \item We randomly select 0.8\% of the lake pixel points from the ground truth, resulting in an image (d). Applying a slight random shift produces an image (e). Next, we generate the unfilled mask prompt using dilate and close operations. To obtain the final filled mask prompt, we identify the contours of the unfilled mask, resulting in an image (f), and perform pixel-filling operations.
\end{itemize}

\begin{table*}
\setlength{\tabcolsep}{3pt}
\fontsize{8.3}{9.0}\selectfont
\centering
\caption{Ablation studies on the different combinations of prompts. CP, RP, FK, and UFK represent three center points, three random points, a filled mask, and an unfilled mask, respectively.}
\begin{tabular}{cc|cccccccccccccccccc}
\hline \hline
\multicolumn{2}{c|}{ID}                           & baseline & 1          & 2              & 3          & 4          & 5          & 6          & 7          & 8          & 9          & 10         & 11         & 12         & 13         & 14         & 15         & 16         & 17         \\ \hline \hline
\multicolumn{1}{c|}{\multirow{2}{*}{Point}} & CP  & -        & \Checkmark & -              & -          & -          & -          & \Checkmark & \Checkmark & \Checkmark & -          & -          & -          & -          & -          & \Checkmark & \Checkmark & -          & -          \\
\multicolumn{1}{c|}{}                       & RP  & -        & -          & \Checkmark     & -          & -          & -          & -          & -          & -          & \Checkmark & \Checkmark & \Checkmark & -          & -          & -          & -          & \Checkmark & \Checkmark \\ \cline{1-2}
\multicolumn{2}{c|}{Box}                          & -        & -          & -              & \Checkmark & -          & -          & \Checkmark & -          & -          & \Checkmark & -          & -          & \Checkmark & \Checkmark & \Checkmark & \Checkmark & \Checkmark & \Checkmark \\ \cline{1-2}
\multicolumn{1}{c|}{\multirow{2}{*}{Mask}}  & FK  & -        & -          & -              & -          & \Checkmark & -          & -          & \Checkmark & -          & -          & \Checkmark & -          & \Checkmark & -          & \Checkmark & -          & \Checkmark & -          \\
\multicolumn{1}{c|}{}                       & UFK & -        & -          & -              & -          & -          & \Checkmark & -          & -          & \Checkmark & -          & -          & \Checkmark & -          & \Checkmark & -          & \Checkmark & -          & \Checkmark \\ \hline \hline
\multicolumn{2}{c|}{mIoU$\uparrow$}               & 90.86    & 91.31      & 91.53 & 91.17      & 91.14      & 91.13      & 91.12      & 90.89      & 90.93      & 91.11      & 90.90      & 90.91      & 90.91      & 90.94      & 90.88      & 90.85      & 90.91      & 90.90      \\ \hline \hline
\end{tabular}
\label{table:combination_of_prompts}
\end{table*}

\section{LEPrompter: Lake Extraction Prompter}

Inspired by the success of SAM~\cite{SAM}, we propose a two-stage prompt enhancement framework, LEPrompter, to integrate the solid prior prompts from the prompt dataset into lake extraction models, which is designed using a combination of prompt-based and prompt-free training during the training stage and prompt-free inference during the inference stage, as depicted in Fig. \ref{TeacherAndStudernt}. LEPrompter consists of a lightweight prompt encoder and decoder with only 1.23M learnable parameters and 0.95G Flops. The architecture of LEPrompter is shown in Fig.~\ref{Overall_architecture} (a). 

 


\subsection{Prompt Encoder}
The function of the prompt encoder is to extract prompt tokens to assist in training the lake extraction model. It processes points and a single box to obtain sparse prompt tokens (SPT), while the mask is processed to obtain a dense prompt token (DPT). A point is represented as the sum of its positional encoding~\cite{PositionalEncoding} and a learned embedding. In contrast to SAM, which processes only 1 point per interaction, we extend the maximum number of points to 9. An embedding pair represents a box: the positional encoding of its top-left corner with a learned embedding for the "top-left corner" and a similar structure for the "bottom-right corner." The DPT is achieved by downscaling using a depth-wise separable convolution~\cite{Depth-wiseConvolutionLayer}, followed by a GELU activation function~\cite{GELUS}. It corresponds spatially to the image embedding $\mathbf{F} \in \mathbb{R} ^{ C \times \frac{H}{4} \times \frac{W}{4}}$, which is obtained by upsampling the feature map from the last encoder layer of the vision image encoder (VIE). If there is no point, box or mask prompt, the prompt encoder outputs a learned embedding representing "no point", "no box", or "no mask." The prompt encoder can be expressed mathematically as:
\begin{equation}  
<\mathbf {O}_s, \mathbf {O}_d> = \mathrm {PE}(\mathbf {Q};<\mathbf {P}_p,\mathbf {P}_b,\mathbf {P}_m>),
\label{Eq:1}
\end{equation}
where $\mathbf {Q}$ is the learnable queries, $\mathbf{P_{p}}, \mathbf{P_{b}}, \mathbf{P_{m}} \in \mathbb{R} ^{ 1 \times H \times W}$ represent the point, box, and mask prompts respectively. $\mathbf{O_{s}} \in \mathbb{R} ^{ N_{tokens} \times C} $ and $\mathbf{O_{d}} \in \mathbb{R} ^{ C \times \frac{H}{4} \times \frac{W}{4}}$ represent the SPT and DPT, where $N_{tokens}\in [1, 12]$. $\mathrm {PE}$ represents the prompt encoder.

\begin{table*}
\small
\centering
\fontsize{8.9}{6.0}\selectfont
\setlength{\tabcolsep}{4.3pt}
\caption{Quantitative comparison of four methods w/o and w/ our proposed approach on the SW, QTPL, CVC-ClinicDB and ISIC2018 datasets. \#P and \#F denote parameters and GFlops with an image size of 256 × 256 in the prompt-based stage.}
\begin{tabular}{@{}cccccccccccccccc@{}}
\toprule
\multirow{2}{*}{Method}     & \multirow{2}{*}{Ours} & \multirow{2}{*}{\#P$\downarrow$} & \multirow{2}{*}{\#F$\downarrow$} & \multicolumn{3}{c}{SW}                            & \multicolumn{3}{c}{QTPL}                         & \multicolumn{3}{c}{CVC-ClinicDB}                 & \multicolumn{3}{c}{ISIC2018}                     \\ \cmidrule(l){5-7} \cmidrule(l){8-10} \cmidrule(l){11-13} \cmidrule(l){14-16} 
                            &                       &                                  &                                  & OA$\uparrow$   & F1$\uparrow$    & mIoU$\uparrow$ & OA$\uparrow$   & F1$\uparrow$   & mIoU$\uparrow$ & OA$\uparrow$   & F1$\uparrow$   & mIoU$\uparrow$ & OA$\uparrow$   & F1$\uparrow$   & mIoU$\uparrow$ \\ \midrule
\multirow{2}{*}{SegFormer~\cite{SegFormer}}  & w/o                   & 3.72                             & 1.59                             & 95.58          & 93.65           & 90.75          & 98.66          & 98.62          & 97.27          & 98.98          & 97.02              & 94.33          & 95.65          & 93.47          & 87.98          \\
                            & w/                     & 5.76                                  & 3.27                                 & \textbf{95.91} & \textbf{95.49}  & \textbf{91.40} & \textbf{98.69} & \textbf{98.64} & \textbf{97.33} & \textbf{99.12} & \textbf{97.43} & \textbf{95.06} & \textbf{96.15} & \textbf{94.30} & \textbf{89.39} \\ \midrule
\multirow{2}{*}{PoolFormer~\cite{PoolFormer}} & w/o                   & 15.64                            & 7.67                             & 94.65          & 94.09           & 88.90          & 98.59          & 98.54          & 97.13          & 98.83          & 96.56              & 93.50          & 95.59          & 93.38          & 87.83          \\
                            & w/                     & 17.12                                 & 7.68                                 & \textbf{95.12} & \textbf{94.62}  & \textbf{89.84} & \textbf{98.62} & \textbf{98.58} & \textbf{97.20} & \textbf{98.97} & \textbf{97.00} & \textbf{94.29} & \textbf{95.77} & \textbf{93.68} & \textbf{88.32} \\ \midrule
\multirow{2}{*}{SegNeXt~\cite{SegNeXt}}    & w/o                   & 4.26                             & 1.55                             & 95.50          & 95.05           & 90.61          & 98.60          & 98.55          & 97.15          & 98.87          & 96.73              & 93.79          & 95.63          & 93.47          & 87.97          \\
                            & w/                     & 6.13                                 & 1.78                                 & \textbf{95.59} & \textbf{95.14}  & \textbf{90.77} & \textbf{98.65} & \textbf{98.60} & \textbf{97.24} & \textbf{99.05} & \textbf{97.22} & \textbf{94.69} & \textbf{95.89} & \textbf{93.81} & \textbf{88.55} \\ \midrule
\multirow{2}{*}{LEFormer~\cite{leformer}}   & w/o                   & 3.61                             & 1.27                             & 95.63          & 95.19           & 90.86          & 98.74          & 98.69          & 97.42          & 98.93          & 96.91              & 94.11          & 95.69          & 93.64          & 88.26          \\
                            & w/                     & 4.84                             & 2.22                             & \textbf{95.97} & \textbf{95.56} & \textbf{91.53} & \textbf{98.75} & \textbf{98.70} & \textbf{97.44} & \textbf{99.13} & \textbf{97.48} & \textbf{95.15} & \textbf{96.33} & \textbf{94.55} & \textbf{89.82} \\ \bottomrule
\end{tabular}
\label{table:results_control_experiment}
\end{table*}

\begin{figure}[t]
\centering
\begin{tabular}{c}
\includegraphics[width=0.85\linewidth]{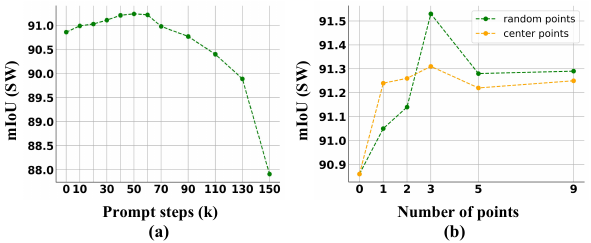} 
\end{tabular}
\caption{(a) Influence of the prompt-based steps. (b) Influence of the type and number of prompt points on our approach.}
\label{Ablation_studies}
\end{figure}

\begin{figure}[t]
\centering
\includegraphics[width=0.38\textwidth]{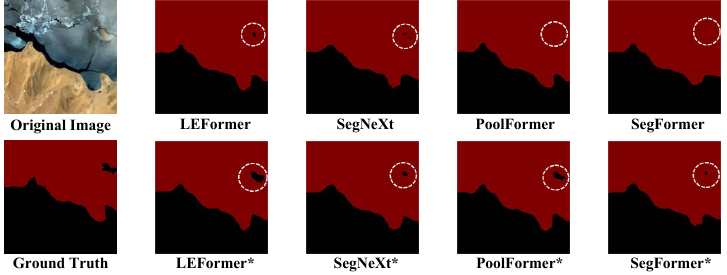} 
\caption{Visualization results on the SW dataset for lake extraction. The white circles highlight noticeable differences. $\mathrm{*}$ denotes w/ our proposed benchmark and baseline.}
\label{controlexperiment}
\end{figure}

\subsection{Prompt Decoder}
We extend the prompt decoder based on SAM to combine one or even three types of prompt tokens from the prompt encoder and the image embedding from the VIE per interaction to generate the final mask by calculating self- and cross-attention. To reduce the complexity of the self- and cross-attention, we employ the sequence reduction process inspired by PVT~\cite{PVT}. Our prompt decoder design, shown in Fig. \ref{Overall_architecture} (b), consists of two Image-Prompt Transformer Blocks (IPTB). Each IPTB performs four steps: (1) efficient self-attention on the tokens; (2) efficient cross-attention from tokens (as queries) to the image embedding; (3) a point-wise MLP updates each token; (4) efficient cross-attention from the image embedding (as queries) to tokens.
This last step updates the image embedding with prompt information while each operation has a residual connection~\cite{ResNet} and a layer normalization. The next IPTB takes the updated $\mathbf{O_s}$ and $\mathbf{O_d}$ from the previous layer and outputs $\mathbf{O_d}$ as an output token, as follows:
\begin{equation}  
\mathbf {O_d} = \mathrm {Conv{(\mathbf{F}+\mathbf {O_d})}},
\label{Eq:2}
\end{equation}
\begin{equation}  
<\mathbf{O_s},\mathbf{O_d}>=\mathrm{IPTB(\mathbf{O_s},\mathbf {O_d})}.
\label{Eq:3}
\end{equation}
The output token is then upsampled and concatenated with other image embeddings from the VIE and then fed to the vision image decoder for the final prediction of the lake mask.

\section{Experiments}

\subsection{Experimental Settings}
\textbf{Implementation Details}. In this work, we train all models on a single Tesla V100 GPU using the \textit{MMSegmentation}~\cite{mmseg2020} codebase for 160K iterations on the SW and QTPL datasets with our benchmark. To assess the generalization capability of our proposed approach, we conduct experiments on two additional binary medical image segmentation datasets: CVC-ClinicDB (CVC)~\cite{bernal2015wm} and ISIC2018~\cite{HAM10000} with an image size of 288 × 384 and 384 × 384, respectively. We randomly divide these datasets into training and testing sets with a 9:1 ratio. We apply data augmentation techniques, such as random resizing and horizontal flipping, using the AdamW optimizer and the cross-entropy loss function with batch size 16. The initial learning rate and weight decay are set to $6\times10^{-5}$ and 0.01.
\\
\textbf{Evaluation Metrics}. 
We evaluate our approach against four methods~\cite{SegFormer, PoolFormer, SegNeXt, leformer}, using metrics such as overall accuracy (OA), F1, and mean Intersection over Union (mIoU) to measure accuracy. We assess efficiency using parameters (Params, M) and floating-point operations per second (Flops, G).

\subsection{Ablation Studies}
We select the SW dataset to evaluate our proposed ablation study method, as the SW dataset presents more challenges due to higher pixel complexity, interclass heterogeneity and complex background information. To evaluate our proposed approach's effectiveness more rigorously, we conduct ablation studies on the SW dataset using LEFormer with our approach.
\\
\textbf{Influence of the prompt-based steps}. 
Fig. \ref{Ablation_studies} (a) shows the model's mIoU improves with increasing prompt-based training steps, peaking at 91.24\% at 50k steps using a single center point prompt. However, in the prompt-free inference phase, excessive training steps may cause over-reliance on prompts, decreasing mIoU. Overall, using our approach appropriately enhances accuracy. We find 50k steps optimal for prompt-based training and apply it in subsequent experiments.
\\
\textbf{Influence of the type and number of prompt points}. 
We evaluate the impact of point prompt types (center vs random points) and prompt point numbers from 0 to 9, as shown in Fig. \ref{Ablation_studies} (b). The model's mIoU increases with more points, peaking at three points, then slightly decreases. At three points, random points yield the highest mIoU of 91.53\%. Thus, we select three random points for subsequent experiments.
\\
\textbf{Influence of the combination of prompts}. 
We examine five prompts with 17 combinations on our benchmark, as shown in Table \ref{table:combination_of_prompts}. Results reveal that most prompt combinations positively affect accuracy, but their effectiveness diminishes with increasing prompt numbers, consistent with the previous experiment. Thus, we select only three random points as the optimal combination of prompts for subsequent experiments.

\subsection{Comparison with State-of-the-Art Methods}
We evaluate our approach against four methods~\cite{SegFormer, PoolFormer, SegNeXt, leformer}. Quantitative results on SW, QTPL, CVC, and ISIC2018 datasets are shown in Table \ref{table:results_control_experiment}. Fig. \ref{controlexperiment} illustrates visualization results for the SW dataset. 

Table \ref{table:results_control_experiment} shows the significant accuracy improvement achieved by our approach. Applying our approach to LEFormer yields mIoU values of 91.53\%, 97.44\%, 95.15\% and 89.82\% on the SW, QTPL, CVC and ISIC2018 datasets, respectively, with improvements of 0.67\%, 0.02\%, 1.04\% and 1.56\%, and only increases of 1.23M parameters and 0.95G Flops in the prompt-based stage. We use only the prompt-free stage during inference without extra parameters and GFlops, preventing higher hardware costs. Overall, our approach is a superior auxiliary framework for lake extraction tasks.

\section{Conclusion}

In this work, we propose three types of prompt datasets (point, box, and mask) and a two-stage prompt enhancement framework called LEPrompter for lake extraction. These aim to improve accuracy without increasing learning difficulty or incorporating excessive noise information. Our benchmark is created by applying morphological operations to the ground truth in the original dataset. Our baseline consists of a lightweight prompt encoder and decoder, which can be easily integrated into existing lake extraction methods to improve accuracy. Experimental results show that our proposed approach significantly enhances the accuracy of automated lake extraction on two widely used datasets. We believe that this study will facilitate and inspire further research in the field of lake extraction.


\section{Acknowledgements}
This work is supported in part by the Natural Science Foundation of China under Grant No. 62222606 and 62076238.

\bibliographystyle{IEEEbib}
\bibliography{icme2023template}

\end{document}